\documentclass{bmvc2k}

\usepackage{bbm}
\usepackage{amssymb}
\usepackage[ruled,linesnumbered]{algorithm2e}
\usepackage{enumitem}
\usepackage[title]{appendix}
\usepackage{wrapfig}
\usepackage{comment}

\newcommand{\tableCellHeight}{1}
\newcommand{\tabstyle}[1]{
  \setlength{\tabcolsep}{#1}
  \renewcommand{\arraystretch}{\tableCellHeight}
  \centering
}

\newcounter{ablmodel} 
\newcommand{\ablmodel}{\refstepcounter{ablmodel}\theablmodel}

\newcommand{\alg}{BORT$^2$}
\newcommand{\keypoint}[1]{\vspace{0.1cm}\noindent\emph{#1}} 

\title{Robust Target Training for Multi-Source Domain Adaptation}

\addauthor{Zhongying Deng}{z.deng@surrey.ac.uk}{1, 2}
\addauthor{Da Li}{dali.academic@gmail.com}{3}
\addauthor{Yi-Zhe Song}{ y.song@surrey.ac.uk}{1,2}
\addauthor{Tao Xiang}{t.xiang@surrey.ac.uk}{1, 2}

\addinstitution{
 University of Surrey\\
 Guildford, UK
}
\addinstitution{
 iFlyTek-Surrey Joint Research Center on Artificial Intelligence
}
\addinstitution{
 Samsung AI Center\\
 Cambridge, UK
}

\runninghead{Deng, Li, Song, Xiang}{Robust Target Training}

\begin{document}

\maketitle

\begin{abstract}
Given multiple labeled source domains and a single target domain, most existing multi-source domain adaptation (MSDA) models are trained on data from all domains jointly in one step. Such an one-step approach limits their ability to adapt to the target domain. This is  because the training set is dominated by the more numerous and labeled source domain data. The source-domain-bias can potentially be alleviated by introducing a second training step, where the model is fine-tuned with the unlabeled target domain data only using pseudo labels as supervision.  However, the pseudo labels are inevitably noisy and when used unchecked can negatively impact the model performance. To address this problem, we propose a novel Bi-level Optimization based Robust Target Training (\alg{}) method for MSDA. 
Given any existing fully-trained one-step MSDA model, \alg{} turns it to a labeling function to generate pseudo-labels for the target data and trains a target model using pseudo-labeled target data only. Crucially, the target model is a stochastic CNN  which is designed to be intrinsically robust against label noise generated by the labeling function. Such a stochastic CNN models each target instance feature as a Gaussian distribution with an entropy maximization regularizer deployed to measure the label uncertainty, which is further exploited to alleviate the negative impact of noisy pseudo labels.  Training the labeling function and the target model poses a nested bi-level optimization problem, for which we formulate an elegant solution based on implicit differentiation. 
Extensive experiments demonstrate that our proposed method achieves the state of the art performance on three MSDA benchmarks, including the large-scale DomainNet dataset. Our code will be available at \url{https://github.com/Zhongying-Deng/BORT2}
\end{abstract}

\section{Introduction}
\label{sec:intro}

Deep convolutional neural networks (CNNs) have advanced significantly in the past decade. In particular, when trained with a large quantity of annotated data~\cite{imagenet_cvpr09}, CNNs have achieved remarkable performance gains over conventional non-CNN-based methods in almost all computer vision tasks, including image classification~\cite{simonyan2014very,szegedy2015going,Deep2016He_resnet,hu2018squeeze}, semantic segmentation~\cite{long2015fully} and object detection~\cite{ren2015faster}. However, this exceptional performance relies on the I.I.D. assumption that the training and test data come from the same underlying distribution independently. When a trained model is applied to data from a different distribution to the training set, its performance often drops significantly. This issue is known as  domain shift~\cite{ben2010theory}, and domain adaptation methods are developed  to address it. A variety of unsupervised domain adaptation (UDA) methods have been proposed~\cite{gretton2012kernel,long2015learning,long2016unsupervised,tzeng2014deep,bhushan2018deepjdot,balaji2019normalized,2018Deep}. Early UDA studies have been focused on the single-source setting~\cite{gretton2012kernel,syn_digits,tzeng2017adversarial}, i.e., adapting a model trained on a single labeled source domain to an unlabeled target domain. Nonetheless, when annotated data collected from multiple source domains are available, training with multiple source domains is expected to help. Therefore, the multi-source domain adaptation (MSDA) setting has received increasing attention since it was first  introduced in~\cite{Peng_2019_ICCV}.

\begin{figure}[t]
    \centering
    \includegraphics[trim=30 100 80 14, clip, width=0.93\columnwidth]{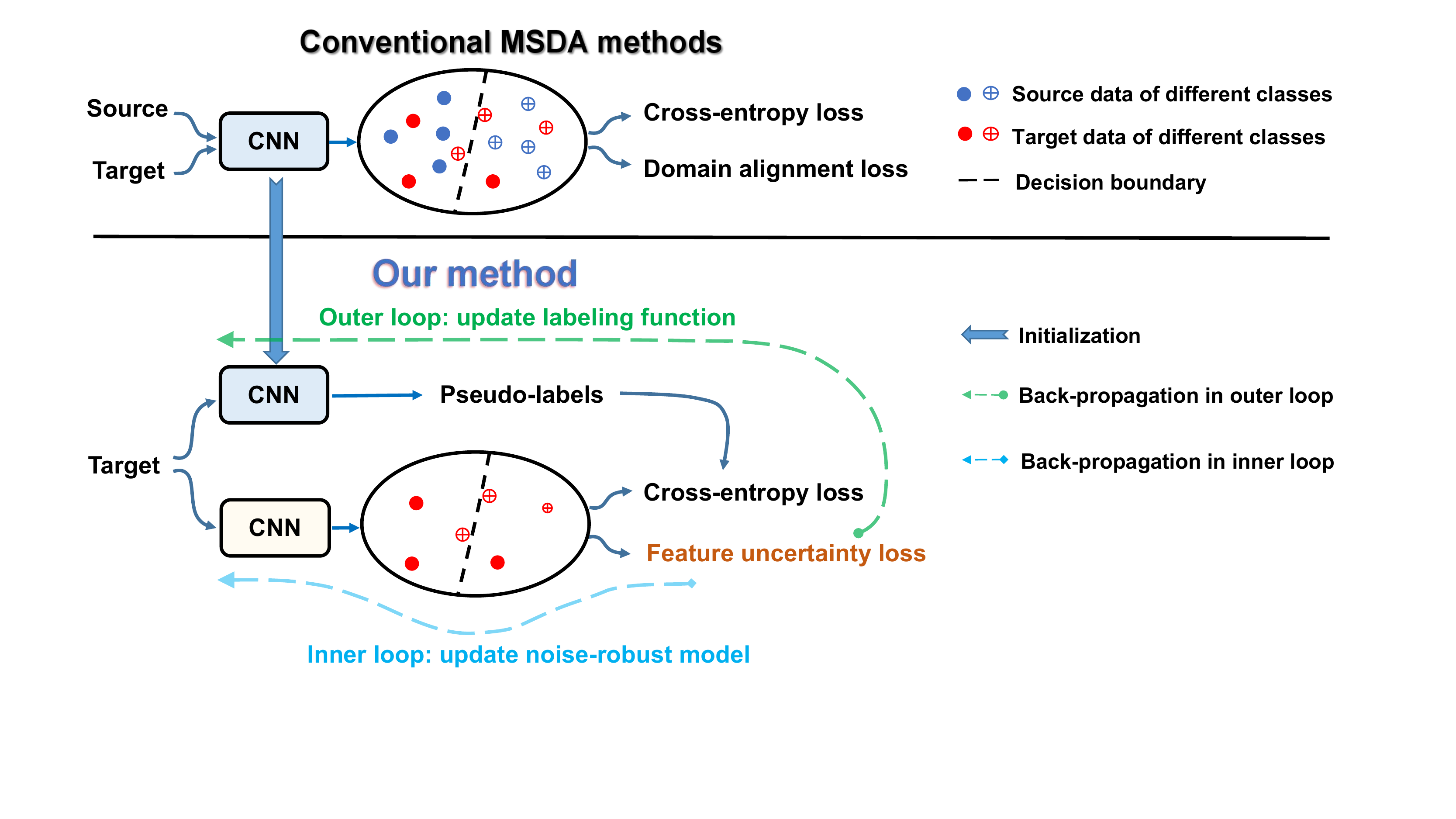}
    \vspace{-.2cm}
    \caption{Our method vs.~conventional MSDA methods. Top: Conventional MSDA models are trained in one step using all domains aiming to extract domain-agnostic features. Bottom: Our method adds a second step training  using the target domain data only. Concretely, the first-step model is fine-tuned to become a labeling function providing supervision for the final target MSDA model (yellow). A stochastic CNN layer is introduced in the final model to make it robust against label noise in the pseudo labels produced by the labeling function on the target data. Both CNNs (labeling function and final model) are learned jointly as a bi-level optimization problem consisting of an inner and outer loop, which is solved using implicit differentiation.}
    \label{fig:intro}

\end{figure}

Most MSDA methods~\cite{2018Deep,Peng_2019_ICCV,wang2020learning,zhou2020domain} adopt an one-step training strategy. As shown in Figure~\ref{fig:intro}, they learn models with a shared backbone to extract domain-agnostic features. In this way, different domains can be aligned in a common feature space. 
However, completely aligning all the domains in one space is extremely difficult and sometimes even counter-productive~\cite{Peng_2019_ICCV}. This is because an one-step MSDA is prone to be biased to the source domains. In particular, since the source domains data are typically in larger quantity (multiple sources vs. one target) and are of higher quality (labeled vs. unlabeled), the one-step trained model would naturally favor the source domains. For instance, it has been observed that the batch norm statistics in a learned MSDA model can be highly source-domain biased~\cite{chang2019domain,mancini2018boosting}. Since a MSDA model is only intended to be used in the target domain, such a bias thus must be addressed.

A naive way to alleviate this source-domain-bias is to introduce a second training step using the unlabeled target domain data only. Concretely, given an one-step MSDA model fully trained using both source and target domain data, the model is fine-tuned in the second step with the target domain data only. Since the target data are unlabeled, a self-training strategy is required, e.g., one can use the pseudo labels generated by the current model for the second-step training in an iterative fashion. Indeed,  we find empirically that given any existing one-step MSDA model, adding a simple pseudo-label based second step training consistently brings a boost to its performance. 

Though such a naive two-step approach can alleviate the source-domain-bias, it brings about another source of bias, i.e., the bias toward erroneous pseudo labels. More specifically, a well-trained first-step MSDA model would not be able to label all target domain data correctly. Otherwise, no second-step adaption is necessary in the first place. These noisy labels, once used directly as supervision, can amplify/re-enforce their bias through the iterations. Simply introducing a threshold to use the model confidence as a pseudo label quality measure can help to a certain extent. But again if we can fully trust the current model to tell us which label is correct, we perhaps do not need the second-step model adaption to start with.

In this work, we propose a novel bi-level optimization based robust target training (\alg{}) method for two-step MSDA (see Figure~\ref{fig:intro}). In the first step, an existing one-step MSDA model is adopted and full-trained on both source and target domains. In the second step, \alg{} uses it as a labeling function to generate pseudo-labels for the target domain data. The model is then trained using the pseudo-labeled target data only. 

We introduce two novel designs to tackle the pseudo-label noise bias.  First, the target model is  designed to be robust against any noisy labels generated by the labeling function.
Specifically, we introduce a stochastic CNN layer in the target model  which models each target instance feature as a Gaussian distribution, consisting of a data dependent mean and variance. We then employ an entropy maximization loss to learn different feature uncertainties (i.e., variances caused by label noise) of different instances as per~\cite{yu2019robust,yu2021simple}.  With this uncertainty measure built in, it is now possible for the target model to identify and subsequently reduces the impact of the noisy labels on model training.

Second, we propose to train both the labeling function and the target model alternatively in a bi-level optimization with an efficient implicit differentiation based solution. That is, the first step (labeling function) and second step (target model) training becomes the outer and inner loops of a nested optimization that alternates between the two steps/loops. In this way, the  labeling function can also be improved to produce less noise.  
However, solving this bi-level optimization problem is non-trivial for two reasons. (a)  The labeling function, a deep CNN itself can now be viewed as a set of `hyper-parameters' for the target stochastic CNN model.  Nevertheless, `hyper-parameter' optimization~\cite{lorraine2020optimizing} typically requires a proper validation set for the outer loop learning objective. In our case, the target domain data is only pseudo labeled with noise, which may harm the optimization when directly used in a validation set. Our solution is to take advantage of the intrinsic uncertainty measure of our stochastic CNN to provide the outer loop learning signal. 
Concretely, in the inner loop we update the target model using the pseudo labels generated by the labeling function. We employ Gumbel-softmax~\cite{jang2016categorical} here when generating the pseudo labels to enable the differentiation of the labeling function. The outer loop computes the predicted feature entropy (uncertainty) of the current training (mini-batch) data using optimized target model in the inner loop. Given that smaller feature uncertainty usually implies an higher probability of accurate labels~\cite{yu2019robust}, the predicted feature uncertainty is minimized to help optimize the labeling function. (b) The hyper-parameters in our cases are the model parameters of a deep CNN, so are in the order of millions thus posing problems for gradient propagation. To overcome this challenge, we use the Neumann series based implicit function theorem~\cite{lorraine2020optimizing} in our bi-level optimization to avoid the computational overload of caching the inner loop optimization trajectories, while maintaining the model convergence in the inner loop optimization. 

We make the following  contributions:
(i) We propose to adopt a  two-step training strategy for MSDA to overcome the source-domain-bias and observe empirically that even a naive pseudo-label based two-step approach brings clear performance boost to a variety of existing MSDA models. 
(ii) To deal with the noisy pseudo labels used for the second-step training, we further propose a novel noise robust training method termed \alg{}, which exploits stochastic CNN for robustness against label noise, and bi-level optimization with joint labeling function training. 
(iii) We show that  the proposed \alg{}  is model agnostic and applicable to any base DA methods (verified with six different MSDA methods). State-of-the-art performance is obtained on three popular MSDA benchmarks, including Digit-Five~\cite{zhou2020domain}, PACS~\cite{li2017deeper} and DomainNet~\cite{Peng_2019_ICCV}.

\section{Related Work}
\label{sec:related_work}
\vspace{-.2cm}
\keypoint{Single-Source Domain Adaptation.} Most single source domain adaptation methods alleviate domain shift by aligning feature distributions between the source and target domains. Some works achieve such feature alignment by minimizing different distance measures, such as maximum mean discrepancy (MMD)~\cite{gretton2012kernel,long2015learning} or Kullback-Leibler (KL) divergence~\cite{zhuang2015supervised}. Some other works employ adversarial training, such as the classic domain adversarial training like DANN~\cite{ganin2016domain} and the more  recent prediction discrepancy based feature/classifier adversarial training, e.g., MCD~\cite{2018Maximum}. 
Our method does not aim for source-target feature alignment. Instead, we focus on how to effectively utilize the target domain to train a model without source bias.

\keypoint{Multi-Source Domain Adaptation (MSDA).} MSDA tackls more practical senerio where multiple source domains are available. Most MSDA methods still attempt to align feature distributions of different domains by using a shared backbone~\cite{zhao2018adversarial,2018Deep,Peng_2019_ICCV}. MDAN~\cite{zhao2018adversarial} and DCTN~\cite{2018Deep} exploit domain adversarial training by training multiple domain discriminators for different source-target domain pairs. M$^3$SDA-$\beta$~\cite{Peng_2019_ICCV} introduces the moment-based distribution distance for different domains. CMSS~\cite{yang2020curriculum} learns a curriculum manager for source sample selection to enable better source/target alignment.
LtC-MSDA~\cite{wang2020learning} explores shared class knowledge among domains by constructing a knowledge graph on the class-wise prototypes of different domains, and exploits such knowledge for better inference. DAC-Net~\cite{deng2021domain}, which extracts domain-invariant features by imposing a consistency loss on the distributions of channel attention weights of different domains. 
DRT~\cite{li2021dynamic} turns multiple source domains into a single source domain problem by using a dynamic model and conduct the feature alignment in a single-source fashion.
Since the shared backbone/classifier inevitably introduces source bias, MDDA~\cite{zhao2020multi} and STEM~\cite{nguyen2021stem} adopts different backbones/classifiers for different domains. 
Although multiple backbones can alleviate the source bias, they introduce more parameters, especially when there are multiple source domains in MSDA. 
Different from these single-step MSDA methods, our work takes a different perspective to alleviate the domain shift and propose a two-step training pipeline. Benefiting from the novel noise robust training scheme, our model can be trained on the target domain only, resulting in better performance than those one-step alternatives.

\begin{figure*}[t]
    \centering
    \includegraphics[trim=0 100 30 120, clip, width=\textwidth]{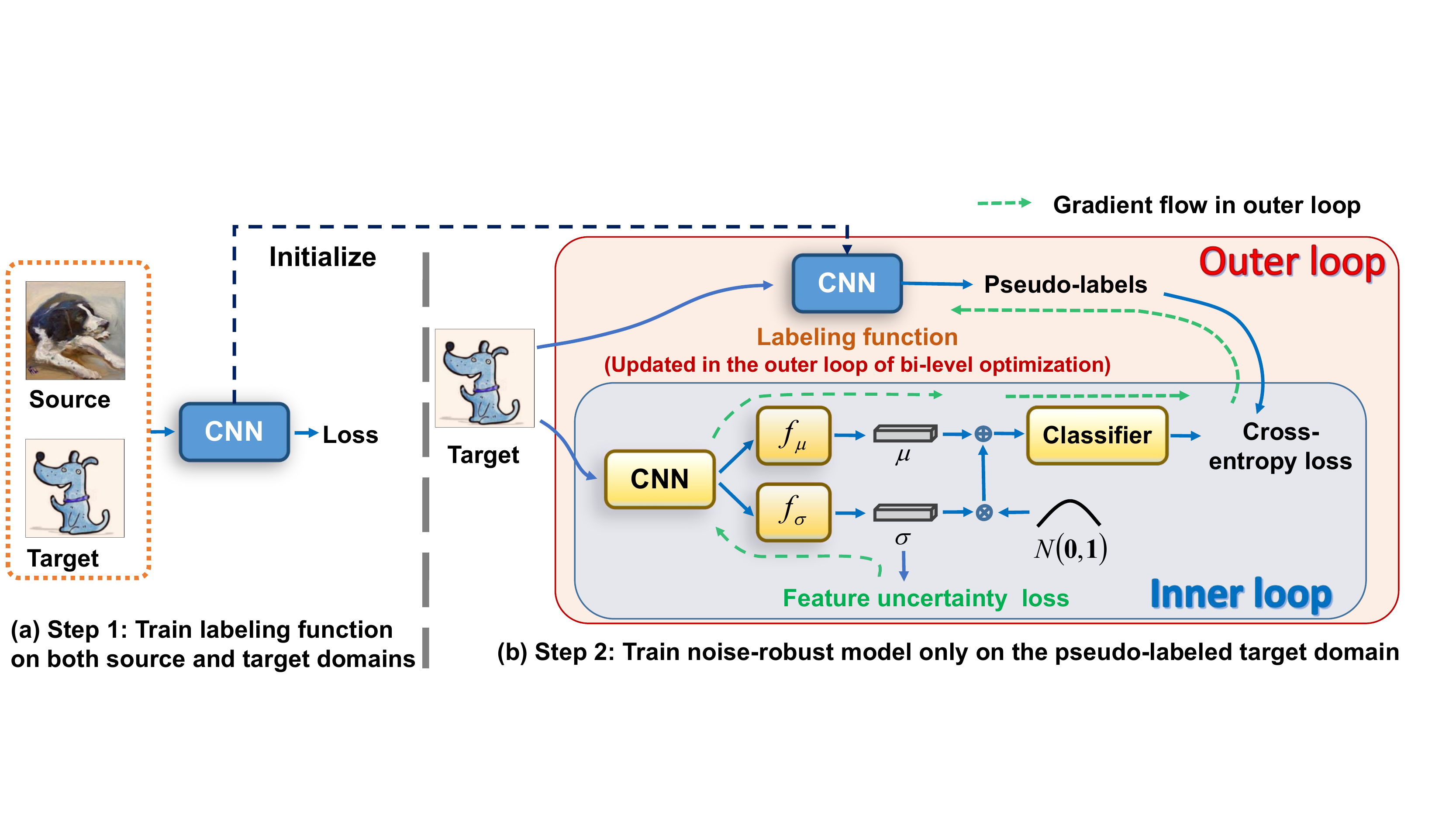}
    \vspace{-.6cm}
    \caption{Overview of our \alg{}. It has two training steps. Step 1 trains a labeling function on both source and target domains. Step 2 trains a target model (the yellow CNN) with only pseudo-labeled target data. 
    The pseudo-labels generated by the labeling function is used for supervised training of the noise-robust target model in the inner loop. The noise-robust model is fed with only target images and outputs predictions for cross-entropy calculation. 
    It models the final feature representation as a Gaussian distribution, with the standard deviation representing the feature uncertainty caused by label noise. An entropy maximization loss is used for learning such feature uncertainty. This entropy loss and the cross-entropy loss are minimized in the inner loop to optimize the noise-robust model. Here, the labeling function is actually a hyper-network for optimizing the noise-robust model. So in the outer loop, we estimate the hyper-parameters of the labeling function for better label quality via bi-level optimization, which is achieved by minimizing the feature uncertainty.}
    \label{fig:overview_ntct}
    \vspace{-0.2cm}
\end{figure*}

\section{Methodology}

In this section, we will introduce the details of our proposed two-step training pipeline for MSDA, including first a naive two-step MSDA method and then our main contribution, the noise robust target model training method \alg{}. The overall training pipeline of \alg{} is shown in Figure~\ref{fig:overview_ntct} and Algorithm ~\ref{alg:ntct}.

\keypoint{Problem Setting.}
This paper focuses on multi-source domain adaptation (MSDA) for image classification. In MSDA, it is typically assumed that there are $K$ labeled source domains $\mathcal{S} = \{ \mathcal{S}_1, ..., \mathcal{S}_K \}$ to adapt to an unlabeled target domain $\mathcal{T}$. Each source domain has $N_{\mathcal{S}_k}$ image and label pairs $\{(x_i^{\mathcal{S}_k}, y_i^{\mathcal{S}_k})\}_{i=1}^{N_{\mathcal{S}_k}}$. The target domain only contains unlabeled images $\mathcal{T}=\{x_{i}^\mathcal{T}\}_{i=1}^{N_\mathcal{T}}$ yet shares the same label space as the source domains. A model is then trained on $\mathcal{D}=\mathcal{S}_1 \cup ...\cup \mathcal{S}_K \cup \mathcal{T}$ jointly and evaluated on a test set of the target domain.

\keypoint{Two-Step Training.} Our two-step training pipeline includes a normal MSDA training step using both source and target domain data,  and a pseudo label based target domain only training step. This pipeline is designed to alleviate the source domain bias.

\subsection{First-Step MSDA Training}
\label{sec:msda}
Let us denote the training model $F_{\theta}$, which is parameterized as $\theta$.
In the first training step of a two-step pipeline,  the MSDA model is learned with the supervision loss from the source domain data and an adaptation loss to align the source and target domains. The overall optimization objective is formulated as
\begin{equation}
    \begin{aligned}
    \underset{\theta}{\arg\min} \sum_{x^s, y^s\sim \mathcal{S}, x^t \sim \mathcal{T}} & \mathcal{L}_{ce}(F_{\theta}(x^s), y^s) + \mathcal{L}_{da}(F_{\theta}(x^s), F_{\theta}(x^t)),
    \end{aligned}
\end{equation}
where, $\mathcal{L}_{ce}$ is a cross entropy loss, and $\mathcal{L}_{da}$ is a domain adaptation loss such as adversarial training~\cite{ganin2016domain} and moment matching~\cite{Peng_2019_ICCV}. This covers most existing MSDA methods. We also introduce FixMatch-CM in Supplementary as a new variant of first-step MSDA  method.

\begin{wrapfigure}[14]{r}{0.54\textwidth}
    \centering
    \vspace{-0.3cm}
    \includegraphics[trim=10 20 5 57, clip, width=0.54\columnwidth]{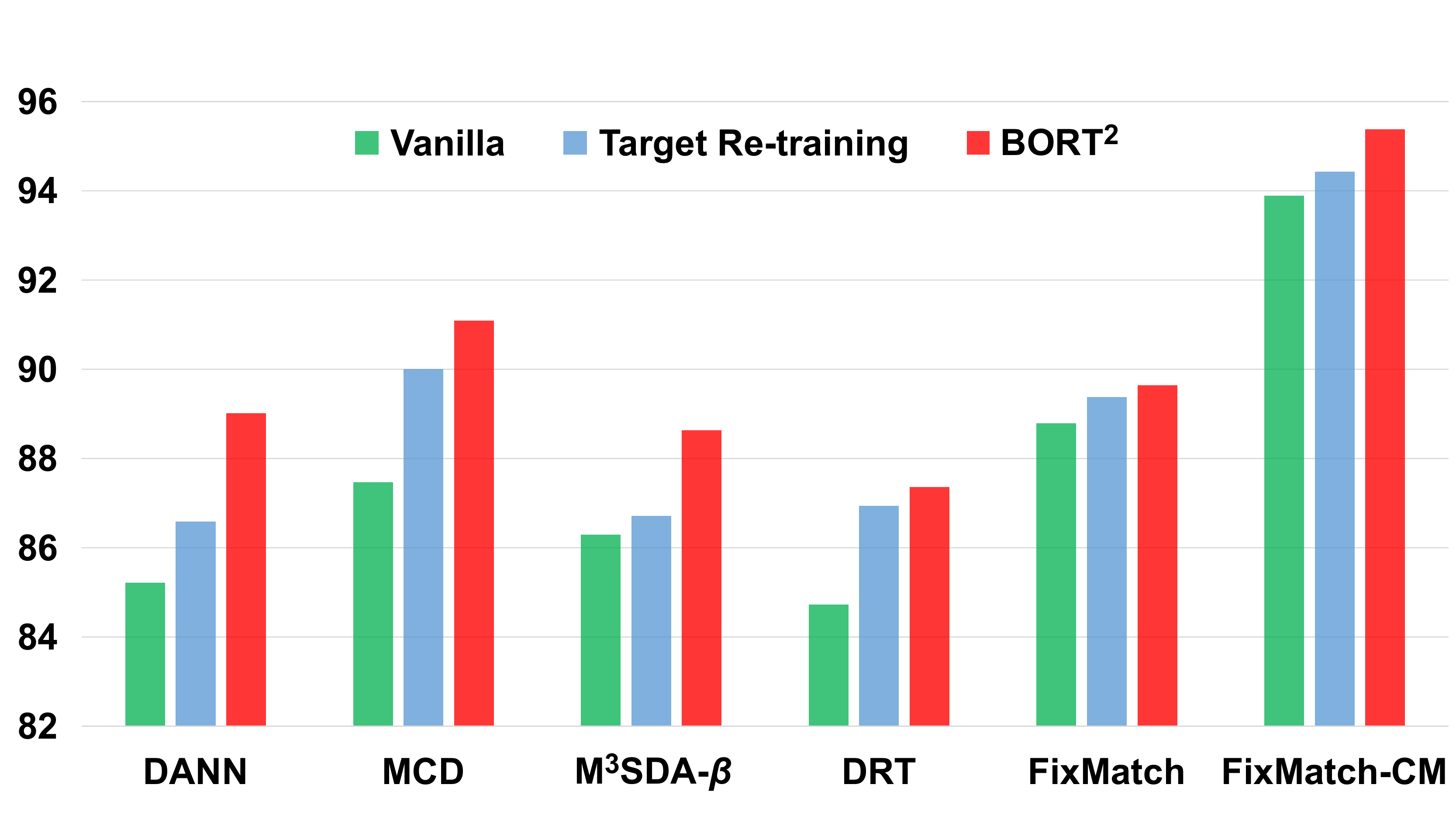}
    \caption{The performance of six one-step MSDA methods (Vanilla) on PACS is improved by a naive second-step target re-training. Our \alg{} further improve the performance significantly.}
    \label{fig:other_sota_with_NTCT}
\end{wrapfigure}

\subsection{Naive Second-Step Training}
As shown in our experiments (see Section \ref{sec:exp}), a simple second step target domain training using pseudo labels can already bring clear improvement on performance, given a variety of existing MSDA models (see Figure~\ref{fig:other_sota_with_NTCT} for a highlight). Let us give some details on this naive training method. 
Note that, in the second training step, there are no labels from the target domain data. Therefore, to train a model on the target domain only, taking a naive approach, we first generate the predictions $p = F_\theta(x), x\sim \mathcal{T}$ using the MSDA model trained in Section~\ref{sec:msda}.
We then convert $p$ to ``hard'' labels: 
\begin{equation}
\label{eq:hardlabel}
    \hat{y}=\arg\max(p). 
\end{equation}
Inspired by FixMatch~\cite{sohn2020fixmatch}, we also put a threshold $\tau$ to select the most confident ``hard'' labels. Meanwhile, we initialize a target domain model $M_{\Psi}$ using $F_\theta$, with $M_{\Psi}$ trained as
\begin{equation}\label{eq:distnet}
\begin{aligned}
        \underset{\Psi}{\arg\min} \ \frac{1}{|\mathcal{T}|}\sum_i \mathbbm{1}(\max(p_i) \geq \tau)\mathcal{L}_{ce}(M_{\Psi}(x_i), \hat{y}_i).
\end{aligned}
\end{equation}

\subsection{Bi-Level Optimization Based Noise-Robust Target Training}
Even after thresholding, the pseudo labels generated for the naive approach is still noisy. Our \alg{} is designed to solve two outstanding problems in the naive approach: 1) how to train a noise-robust model on the pseudo-labeled target domain with label noise. And 2) how to improve the labeling function further to provide higher-quality pseudo-labels. Two mechanisms are formulated in \alg{} to solve these two problems respectively.

\subsubsection{Stochastic Feature Uncertainty Modeling.} Inspired by the noisy-label learning methods in \cite{yu2019robust,yu2021simple}, we introduce stochastic modeling in the fully-trained first-step model $F_{\theta}$ to turn it into a robust final model  ${M}^{}_{\Psi}$ that can cope with the noisy pseudo labels used for supervision. More specifically, we introduce a stochastic layer to the final feature output of $F_{\theta}$. Such a layer models each instance feature $z^{l}_i$ produced by the $l^{th}$ (final) feature layer of $F_{\theta}$ as a Gaussian distribution, i.e. $z^{l}_i\sim N(\mu_i, \sigma_i^2)$, in which $\mu,\sigma$ are generated as
\begin{equation}
    \begin{aligned}
    \mu = f_{\Psi_\mu}(z^{l-1}),\quad \sigma = f_{\Psi_\sigma}(z^{l-1}),
    \end{aligned}
\end{equation}
where $z^{l-1}=f^{l-1}\circ\dots\circ f^{1}(x)=f_{\Psi_0}(x)$, and $f^i$ is a feature layer. $f_{\Psi_\mu}, f_{\Psi_\sigma}$ are the learnable layers that output $\mu, \sigma$. $x$ is the input with a pseudo label $\hat{y}$ sampled from the training set $\{x, \hat{y}\}_{i}$. 
And, a reparameterization trick is employed for enabling the back propagation as $z_i = \mu_i + \sigma_i \cdot \epsilon$, where $\epsilon \sim N(0, \mathbb{I})$.
Then, a classifier $g_{\Psi_1}(.)$ is followed to classify $z_i$. The learning objective formula of the robust final model $M_{\Psi}$ is 
\begin{equation}
\label{eq:loss_inner}
\begin{aligned}
    \underset{\Psi=\{\Psi_0, \Psi_\mu, \Psi_\sigma, \Psi_1\}}{\arg\min}~ \mathcal{L}_{trn} =  \frac{1}{|\mathcal{T}|} \sum_{x_i\sim \mathcal{T}} \mathbbm{1}(\max(p_i) \geq \tau)\mathcal{L}_{ce}(g_{\Psi_1}(z_i), \hat{y}_i)   + \lambda  \mathcal{L}_{ment}(f_{\Psi_\sigma}(f_{\Psi_{0}}(x_i))),
\end{aligned}
\end{equation}
consisting of a cross-entropy loss $\mathcal{L}_{ce}$ and an entropy maximization loss $\mathcal{L}_{ment}(\sigma_{i})= (m - \sum\mathrm{log}(\sigma_{i}))^+$ where $m$ is a margin to bound the uncertainty. During the optimization, the optimizer will choose to assign larger standard deviation to the noisy labels as it will cancel its learning signal out, otherwise the loss will be enlarged significantly~\cite{yu2019robust}. In other words, the model is able to automatically identify those uncertain therefore noisy instance labels and discount their influence on model training.

\subsubsection{Bi-level Optimization of Labeling Function.} In this section, we will introduce how we further improve the labeling function to generate better-quality pseudo labels.
The final model $M_{\Psi}$ is trained with the pseudo labels generated by the first-step model $F_{\theta}$. This means that the trained model is conditioned on the pseudo labels, i.e., the labeling function $F_{\theta}$. Optimizing the function $F_{\theta}$ thus becomes an `hyperparameter' optimization (HO) problem, which can be formulated as
\begin{equation}\label{eq:bilevel}
    \underset{\theta}{\operatorname{argmin}}~ \mathcal{L}_{val} \big( \underset{\Psi}{\operatorname{argmin}} ~\mathcal{L}_{trn} (M_\Psi, \mathcal{T}_{trn} ; F_\theta), \mathcal{T}_{val}  \big),
\end{equation}
where $\theta$ can be regarded as the hyperparemters of model $M_\Psi$. $\mathcal{T}_{trn}$ and $\mathcal{T}_{val}$ are training and validation sets respectively, and $\mathcal{L}_{val}$ is the validation objective minimized to optimize $\theta$.

In this bi-level optimization, the inner loop learning objective $\mathcal{L}_{trn}$ is the same as Eq.~\eqref{eq:loss_inner}, except that the pseudo-label $\hat{y}_i$ is generated by using Gumbel-Softmax~\cite{maddison2016concrete,jang2016categorical} as $\hat{y}_i=\mathrm{GumbelSoftmax}(F_{\theta}(x_i))$ to enable the back-propagation of $F_\theta$ in the outer loop optimization. Note that, typically $\mathcal{T}_{val}$ is a held-out validation set, which is used to compute the validation loss of the \emph{best-response} model\cite{lorraine2020optimizing} $M_\Psi$ to optimize the hyperparameters $\theta$. However, in our case the target domain data is only pseudo-labeled with noise. Directly using a validation set constructed from those noisy pseudo labels will harm the outer loop optimization.

In the entropy maximization in Eq.~\eqref{eq:loss_inner}, we know that the optimizer will choose to assign larger entropy (uncertainty) to the noisy labels. Therefore, the entropy can be explicitly used as a measure of how noisy a predicted label is. That is to say, labels with lower uncertainty are more likely to be accurate labels. Thus, we choose to use the entropy loss as our validation loss in Eq.~\eqref{eq:bilevel}, i.e.
\begin{equation}
\label{eq:loss_outer}
\begin{aligned}
    \mathcal{L}_{val} (M_\Psi^*, \mathcal{T}_{val}) &= \frac{1}{|\mathcal{T}_{val}|}\sum_{x_i \sim \mathcal{T}_{val}}\big(\sum\mathrm{log}(f_{\Psi_\sigma^*}(f_{\Psi^*_0}(x_i))) \big),
\end{aligned}
\end{equation}
where $\Psi^*=\operatorname{argmin}_{\Psi} ~\mathcal{L}_{trn} (M_\Psi, \mathcal{T}_{trn} ; F_\theta)$ is the converged model  in the inner loop under the hyperparameter $\theta$. Note that we use $\mathcal{T}_{val}=\mathcal{T}_{trn}$ here. Our objective is to optimize the labeling function such that the predicted feature uncertainty of \emph{training data} is low when using the generated labels from the labeling function. Therefore, it makes more sense to validate the feature uncertainty of the training set for the sake of optimizing our labelling function.

During the outer optimization, the hypergradient~\cite{lorraine2020optimizing} of $\theta$ is computed as
\begin{equation}
   \frac{ \partial \mathcal{L}_{val} (M_{\Psi^*}) }{\partial \theta} = \frac{ \partial \mathcal{L}_{val}}{\partial M_{\Psi^*}} \frac{ \partial  M_{\Psi^*} }{\partial \theta},
\end{equation}
where $\frac{ \partial \mathcal{L}_{val}}{\partial M_{\Psi^*}}$ can be straightforwardly computed using existing deep learning tools, e.g. PyTorch. $\frac{\partial  M_{\Psi^*} }{\partial \theta}$ can be decomposed into $\frac{\partial  M_{\Psi^*} }{\partial \theta}=-\big[\frac{\partial^2  \mathcal{L}_{trn} }{\partial \Psi\partial \Psi  }\big]^{-1} \times \frac{\partial^2  \mathcal{L}_{trn} }{\partial \Psi\partial \theta  }$ according to Implicit Function Theorem~\cite{lorraine2020optimizing}. Computing the inverse Hessian is not tractable in the high dimensional space. Therefore, we use a recently published Neumann approximation~\cite{lorraine2020optimizing}.

\begin{algorithm}[t]
\caption{Training Procedure of \alg{}}
\label{alg:ntct}
\SetKwInOut{Input}{input}\SetKwInOut{Output}{output}
\Input{ Training data $\{(x_i^{\mathcal{S}_1}, y_i^{\mathcal{S}_1}), ..., (x_i^{\mathcal{S}_K}, y_i^{\mathcal{S}_K})\}_{i=1}^B$, $\{x_i^{\mathcal{T}}\}_{i=1}^B$.}
\Output{ The target domain model $M_{\psi}$.}
   \While{not converge}{
       Update labeling function $F_\theta$ via any existing MSDA methods. 
   }
   \While{not converge} {
      Update target domain model $M_{\psi}$ via Eq. ~\eqref{eq:loss_inner}.\\
      Bi-level optimization for labeling function $F_\theta$:\\
         \Indp Inner loop optimization according to Eq. ~\eqref{eq:loss_inner} with Gumbel-softmax.\\ 
         Outer loop optimization via minimizing Eq. ~\eqref{eq:bilevel} using Neumann approximation~\cite{lorraine2020optimizing}.\\
  }
\end{algorithm}

\section{Experiments}
\label{sec:exp}
We experiment on three popular MSDA datasets, including PACS~\cite{li2017deeper}, Digit-Five, and  DomainNet~\cite{Peng_2019_ICCV}. The experimental setting are provided in Supplementary Material.

\subsection{Comparative Results}

\begin{wraptable}[9]{r}{0.5\textwidth}
    \centering
    \tabstyle{2pt}
    \vspace{-.7cm}
    \caption{MSDA results on PACS. Best results are in bold.}
    \resizebox{0.5\textwidth}{!}{
    \begin{tabular}{l|cccc|c}
    \hline
    \textbf{Method} & \textbf{Art.} & \textbf{Cartoon} & \textbf{Sketch} & \textbf{Photo} & \textbf{Avg.}\\
    \hline 
    Oracle  &99.53 &99.84 &99.53 &99.92 & 99.71\\
    \hline
    Source-only &81.22	&78.54	&72.54	&95.45	&81.94\\
    MDAN~\cite{zhao2018adversarial} &83.54 &82.34 &72.42 &92.91 &82.80\\
    DCTN~\cite{2018Deep}  &84.67 &86.72 &71.84 &95.60 &84.71\\
    M$^3$SDA-$\beta$~\cite{Peng_2019_ICCV}  &84.20 &85.68 &74.62 &94.47 &84.74\\
    MDDA~\cite{zhao2020multi} &86.73 &86.24 &77.56 &93.89 &86.11\\
    LtC-MSDA~\cite{wang2020learning} &90.19 &90.47 &81.53 &97.23 &89.85\\
    DAC-Net~\cite{deng2021domain} &91.39	&91.39	&84.97	&97.93	&91.42 \\
    \hline
    \alg{} (\emph{Ours}) &\textbf{95.02} &\textbf{94.51} &\textbf{93.23} &\textbf{98.74} &\textbf{95.38}\\
    \hline
    \end{tabular}
    }
    \label{tab:sota_pacs}
\end{wraptable}

\paragraph{Competitors}
We compare our method with the following competitors introduced in Section~\ref{sec:related_work}:  DANN~\cite{ganin2016domain}, MCD~\cite{2018Maximum}, MDAN~\cite{zhao2018adversarial}, DCTN~\cite{2018Deep}, M$^3$SDA-$\beta$~\cite{Peng_2019_ICCV},   LtC-MSDA~\cite{wang2020learning}, DAC-Net~\cite{deng2021domain}, CMSS~\cite{yang2020curriculum}, MDDA~\cite{zhao2020multi}, DRT~\cite{li2021dynamic} and STEM~\cite{nguyen2021stem}. Most of these methods try to minimize domain gap via a one-step training, thus can hardly alleviate source-domain-bias.

\paragraph{PACS} From Table~\ref{tab:sota_pacs}, we can see that \alg{} is superior to these competitors on all four transfer tasks, leading to an average accuracy of 3.96\% improvement over other baselines. On some difficult setups, such as Sketch and Art Painting as target domains, \alg{} outperforms the second best method by 8.26\% and 3.63\% respectively. This demonstrate the strong robustness of our \alg{} under large domain shifts.

\vspace{-.5cm}
\paragraph{Digit-Five}
As shown in Table~\ref{tab:sota_digit_five}, \alg{} achieves significant improvement over the previous state-of-the-art methods,  e.g. 4\% better than DRT in average accuracy, 1.4\% than DAC-Net and $\sim$1\% than STEM. In particular, our \alg{} obtains comparable performance to the oracle result, demonstrating the high-quality pseudo-labels generated. On the MNIST-M domain, \alg{} shows biggest improvement over the other competitors (with 3.3\%).

\vspace{-.5cm}
\paragraph{DomainNet}
Table~\ref{tab:sota_domainnet} shows that \alg{} achieves comparable performance with STEM, but does not adopt classifier ensemble strategy as STEM. In addition, \alg{} beats the other competitors considerably, with more than 2.2\% performance gain. On the most challenging target domain Quickdraw, our \alg{} obtains more than 2.0\% improvement over the other methods. This further verifies the effectiveness of \alg{} for addressing large domain shift, thanks to its robust target training.

\begin{table*}[tb]
    \centering
    \caption{MSDA results on Digit-Five. * denotes that standard deviations are not reported in the paper.}
    \resizebox{0.9\textwidth}{!}{
    \begin{tabular}{l|ccccc|c}
    \hline
    \textbf{Method} & \textbf{MNIST} & \textbf{USPS} & \textbf{MNIST-M} & \textbf{SVHN} & \textbf{Synthetic} & \textbf{Avg.}\\
    \hline 
    Oracle & 99.5$\pm$0.03 &99.1$\pm$0.05 &95.0$\pm$0.29 & 90.7$\pm$0.26 & 97.8$\pm$0.02 & 96.4\\
    \hline
    Source-only~\cite{yang2020curriculum}  & 92.3$\pm$0.91 &90.7$\pm$0.54  &63.7$\pm$0.83 &71.5$\pm$0.75 &83.4$\pm$0.79 &80.3\\ 
    DANN~\cite{ganin2016domain} &97.9$\pm$0.83 &93.4$\pm$0.79 &70.8$\pm$0.94 &68.5$\pm$0.85 &87.3$\pm$0.68 &83.6\\ 
    DCTN~\cite{2018Deep} & 96.2$\pm$0.80  &92.8$\pm$0.30 &70.5$\pm$1.20 &77.6$\pm$0.40 &86.8$\pm$0.80 &84.8 \\
    MCD~\cite{2018Maximum}  & 96.2$\pm$0.81 &95.3$\pm$0.74 &72.5$\pm$0.67 &78.8$\pm$0.78 &87.4$\pm$0.65 &86.1\\ 
    M$^3$SDA-$\beta$~\cite{Peng_2019_ICCV}   &  98.4$\pm$0.68 &96.1$\pm$0.81 &72.8$\pm$1.13 &81.3$\pm$0.86 &89.6$\pm$0.56 &87.6\\ 
    CMSS~\cite{yang2020curriculum} &99.0$\pm$0.08 &97.7$\pm$0.13 &75.3$\pm$0.57 &88.4$\pm$0.54 &93.7$\pm$0.21 &90.8 \\ 
    LtC-MSDA~\cite{wang2020learning}   &99.0$\pm$0.40 &98.3$\pm$0.40 &85.6$\pm$0.80 &83.2$\pm$0.60 &93.0$\pm$0.50 &91.8\\
    DRT~\cite{li2021dynamic} &\textbf{99.3}$\pm$0.05 &98.4$\pm$0.12 &81.0$\pm$0.34 &86.7$\pm$0.38 &93.9$\pm$0.34 &91.9\\
    DAC-Net~\cite{deng2021domain} &99.2$\pm$0.03 &\textbf{98.7}$\pm$0.11  &86.0$\pm$0.44  &91.6$\pm$0.16  &97.1$\pm$0.18 & 94.5 \\
    STEM~\cite{nguyen2021stem}\textsuperscript{*} &99.4 &98.4 &89.7 &89.9 &\textbf{97.5} &95.0\\
    \hline
    \alg{} (\emph{Ours}) &98.8$\pm$0.08 &98.4$\pm$0.08 &\textbf{93.0}$\pm$0.06 &\textbf{91.9}$\pm$0.19 &\textbf{97.5}$\pm$0.08 &\textbf{95.9}\\
    \hline
    \end{tabular}
    }
    \label{tab:sota_digit_five}
\end{table*}

\begin{table*}[t]
    \centering
    \caption{MSDA results on DomainNet.}
    \resizebox{\textwidth}{!}{
    \begin{tabular}{l|cccccc|c}
    \hline
        \textbf{Method} & \textbf{Clipart} & \textbf{Infograph} & \textbf{Painting} & \textbf{Quickdraw} & \textbf{Real}  & \textbf{Sketch} & \textbf{Avg.}\\
    \hline %
        Oracle & 79.7$\pm$0.16 &41.0$\pm$0.18 & 71.4$\pm$0.11 &72.6$\pm$0.70 &83.7$\pm$0.13 &70.59$\pm$0.06 &69.8\\
    \hline
        Source-only~\cite{Peng_2019_ICCV} &47.6$\pm$0.52 &13.0$\pm$0.41 & 38.1$\pm$0.45 &13.3$\pm$0.39 &51.9$\pm$0.85 &33.7$\pm$0.54 &32.9\\
        DANN~\cite{ganin2016domain} &45.5$\pm$0.59 &13.1$\pm$0.72 &37.0$\pm$0.69 &13.2$\pm$0.77 &48.9$\pm$0.65 &31.8$\pm$0.62 &32.6\\
        DCTN~\cite{2018Deep} &48.6$\pm$0.73 &23.5$\pm$0.59 &48.8$\pm$0.63 &7.2$\pm$0.46 &53.5$\pm$0.56 &47.3$\pm$0.47 &38.2 \\
        MCD~\cite{2018Maximum} &54.3$\pm$0.64 &22.1$\pm$0.70 &45.7$\pm$0.63 &7.6$\pm$0.49 &58.4$\pm$0.65 &43.5$\pm$0.57 &38.5\\
        M$^3$SDA-$\beta$~\cite{Peng_2019_ICCV} &58.6$\pm$0.53 &26.0$\pm$0.89 &52.3$\pm$0.55 &6.3$\pm$0.58 &62.7$\pm$0.51 &49.5$\pm$0.76 &42.6\\
        CMSS~\cite{yang2020curriculum} &64.2$\pm$0.18 &28.0$\pm$0.20 &53.6$\pm$0.39 &16.0$\pm$0.12 &63.4$\pm$0.21 &53.8$\pm$0.35 &46.5\\
        LtC-MSDA~\cite{wang2020learning} &63.1$\pm$0.50 &28.7$\pm$0.70 &56.1$\pm$0.50 &16.3$\pm$0.50 &66.1$\pm$0.60 &53.8$\pm$0.60 &47.4\\
        DRT~\cite{li2021dynamic} &69.7$\pm$0.24 &\textbf{31.0}$\pm$0.56 &59.5$\pm$0.43 &9.9$\pm$1.03 &68.4$\pm$0.28 &59.4$\pm$0.21 &49.7\\
        DAC-Net~\cite{deng2021domain} &72.5$\pm$0.04	&27.6$\pm$0.10	&57.8$\pm$0.06	  &23.0$\pm$0.14	&66.7$\pm$0.10	&59.5$\pm$0.12	&51.2\\
        STEM~\cite{nguyen2021stem} &72.0 &28.2 &\textbf{61.5} &25.7 &\textbf{72.6} &60.2 &\textbf{53.4}\\
    \hline
        \alg{} (\emph{Ours}) &\textbf{74.0}$\pm$0.04 &29.1$\pm$0.19 &59.6$\pm$0.06 &\textbf{28.0}$\pm$0.02 &69.3$\pm$0.04 &\textbf{60.3} $\pm$0.14 &\textbf{53.4}\\
    \hline
    \end{tabular}
    }
    \label{tab:sota_domainnet}
    \vspace{-0.5cm}
 \end{table*}

\subsection{Further Analysis}

\keypoint{Importance of a Second Step Training.}
We verify the contribution of our proposed robust target training here. From Figure~\ref{fig:other_sota_with_NTCT}, we can see that a simple second step target domain training using pseudo labels improves all six differnet base MSDA methods, resulting in accuracy improvements of 1.37\% on DANN~\cite{ganin2016domain}, 2.53\% on MCD~\cite{2018Maximum}, 0.42\% on M$^3$SDA-$\beta$~\cite{Peng_2019_ICCV}, 2.21\% on DRT~\cite{li2021dynamic} and 0.59\%, 0.54\% on two FixMatch~\cite{sohn2020fixmatch} variants (or see \#\ref{ablmodel:ntct_var_wo_nrm} vs. \#\ref{ablmodel:fm_ms_cm} in Table~\ref{tab:ablation_pacs}). Incorporating our proposed robust training further improves this second step training, with a up to 2.43\% accuracy gain. 

\begin{wraptable}[6]{r}{0.5\textwidth}
    \centering
    \vspace{-0.4cm}
    \caption{Ablation study of \alg{} on PACS.}
    \resizebox{0.5\columnwidth}{!}{
    \begin{tabular}{l|l|cccc|c}
    \hline
    \textbf{\#} &\textbf{Methods} & \textbf{Avg}\\
    \hline
    \ablmodel  \label{ablmodel:ntct_final} & \alg{} & \textbf{95.38}\\
    \ablmodel  \label{ablmodel:ntct_var_wo_biopt} & \alg{} (w/o bi-level optimization) & 94.80\\
    \ablmodel  \label{ablmodel:ntct_var_wo_nrm} & \alg{} (w/o noise-robust model) & 94.43\\
    \ablmodel  \label{ablmodel:fm_ms_cm} & FixMatch-CM &93.89\\
    \hline
    \end{tabular}
    }
    \label{tab:ablation_pacs}
\end{wraptable}

\keypoint{Importance of Optimizing Labeling Function.}
In the second step of \alg{}, we propose to optimize the labeling function by a bi-level optimization. To verify its effectiveness, we remove the outer loop in Eq.~\eqref{eq:loss_outer} from \#\ref{ablmodel:ntct_final} but keep the stochastic modelling. This leads to a model without bi-level optimization, further resulting in a fixed labeling function. From Table~\ref{tab:ablation_pacs} \#\ref{ablmodel:ntct_var_wo_biopt} we can see that without this bi-level optimization, the performance decreases by 0.58\% from \#\ref{ablmodel:ntct_final}. This indeed shows that optimizing the labeling function is helpful to improve the quality of pseudo-labels.

\keypoint{Importance of Noise-Robust Training.}
We further evaluate the noise-robust training used in the second step of \alg{} by replacing the feature uncertainty based stochastic model in \#\ref{ablmodel:ntct_var_wo_biopt} with a vanilla CNN. This leads to a naive second-step training in Table~\ref{tab:ablation_pacs}\#\ref{ablmodel:ntct_var_wo_nrm}. Comparing \#\ref{ablmodel:ntct_var_wo_nrm} with \#\ref{ablmodel:ntct_var_wo_biopt}, we observe a performance drop, suggesting that this stochastic modelling is helpful.

\keypoint{Sensitivity of Hyper-Parameters}
\label{sec:hyper_param}
\begin{figure}[t]
    \centering
    \subfigure{
        \begin{minipage}[t]{0.4\textwidth}
        \includegraphics[trim=5 15 50 5,clip,width=\columnwidth]{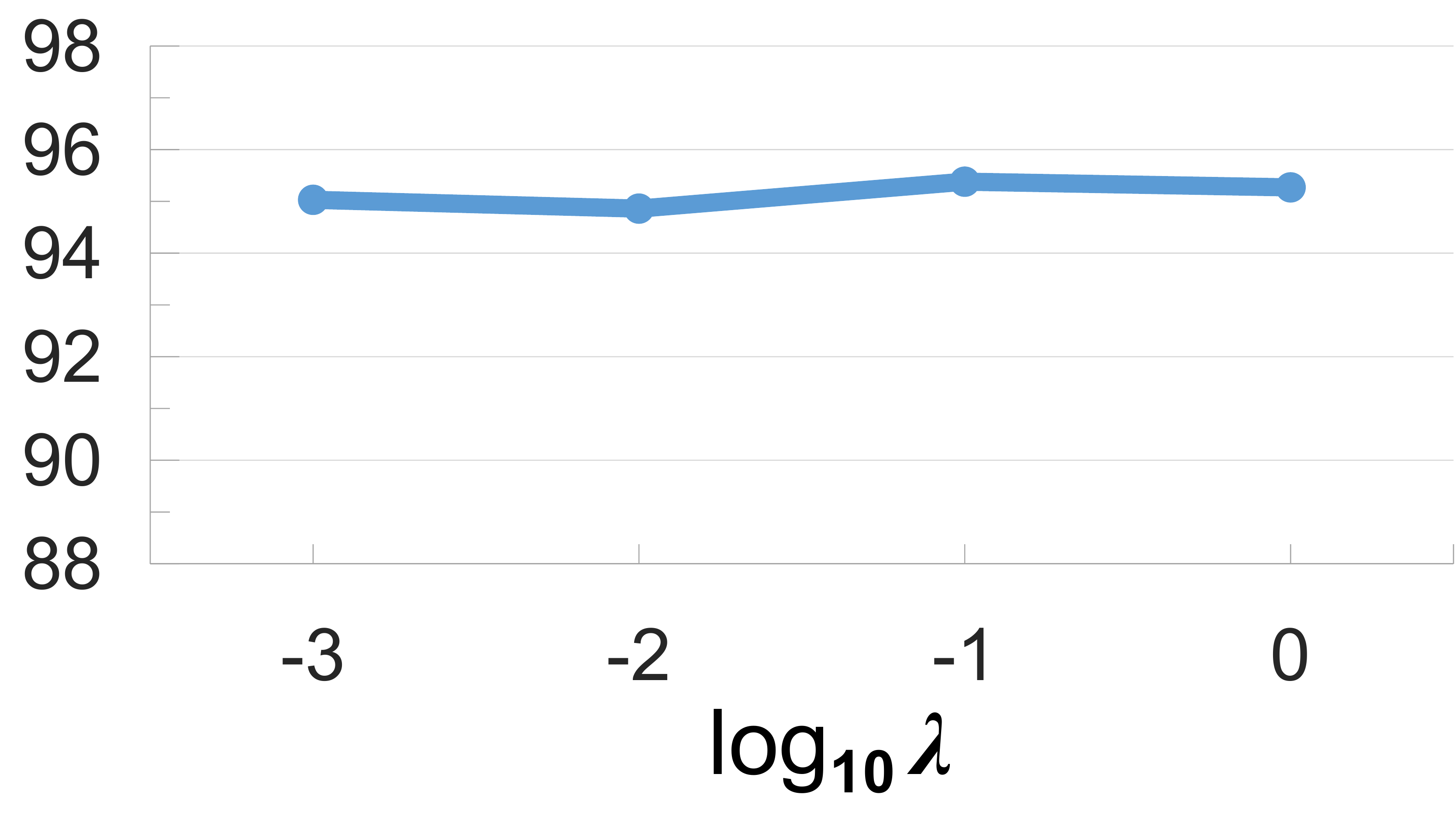}
        \label{fig:lambda}
        \end{minipage}

    }
    ~\hspace{0.2cm}
    \subfigure{
        \begin{minipage}[t]{0.4\textwidth}
        \includegraphics[trim=5 15 50 5,clip,width=\columnwidth]{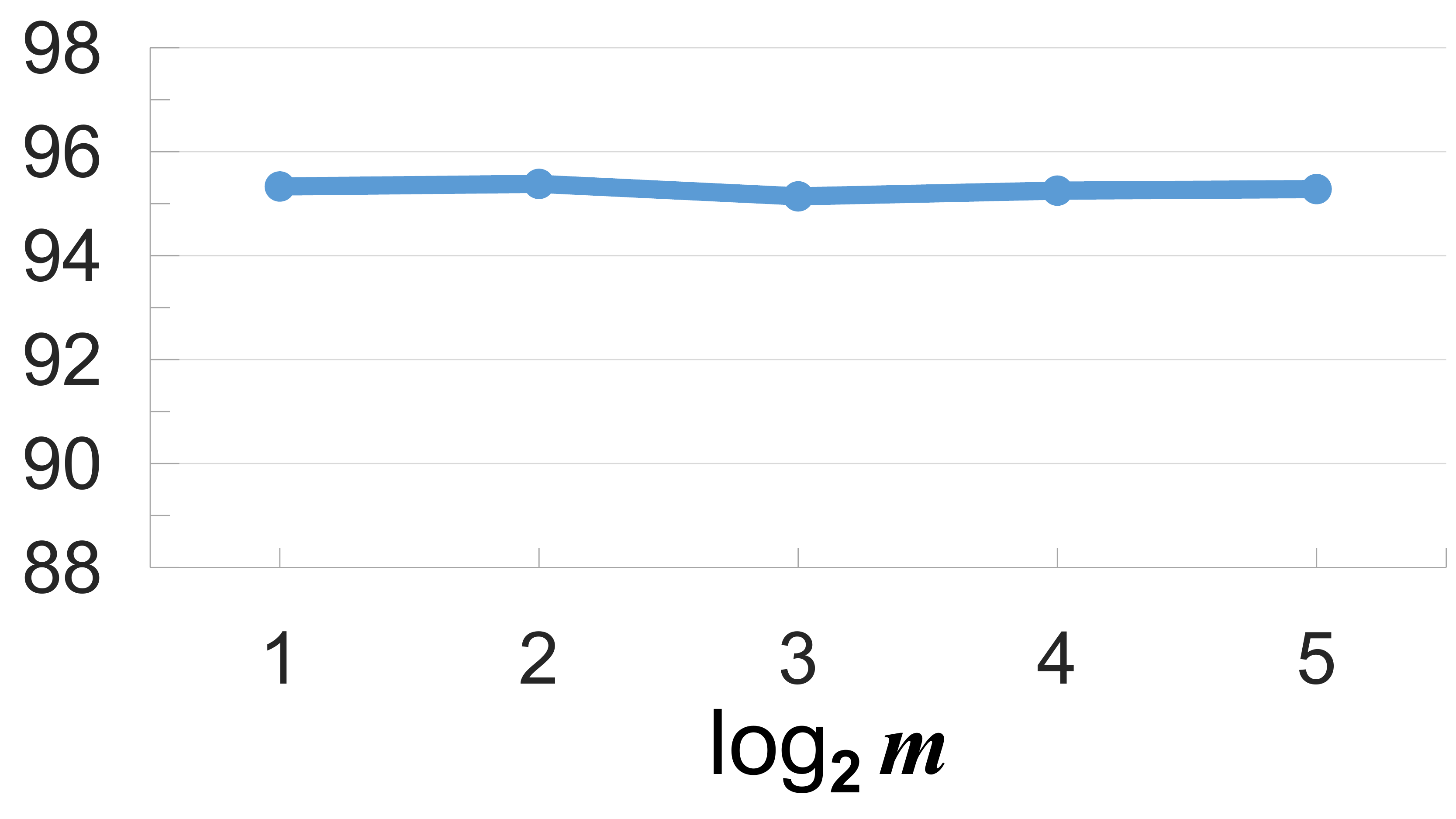}
        \label{fig:threshold}
        \end{minipage}
    }
     \vspace{-0.4cm}
    \caption{Sensitivity of $\lambda$ and $m$ in $L_{ment}$.}
    \label{fig:hyper_param}
    \vspace{-0.4cm}
\end{figure}
Recall that in our proposed \alg{}, we have two hyper-parameters: the weight $\lambda$ and threshold $m$ in the entropy maximization loss $L_{ment}$ (see Eq. (5) in the main paper). We first fix $m$ to 4 and vary $\lambda$ from 0.001 to 1. The results are in Figure~\ref{fig:hyper_param} (left panel). 
It is clear that the performance is generally stable, and the best performance of 95.38\% is obtained at $\lambda=0.1$ ( i.e., $\log_{10} \lambda=-1$). We then set $\lambda$ to 0.1 and adjust $m$ from 2 to 32. The results (right panel) show that the performance is also insensitive to $m$, with the best accuracy achieved at $m=4$ (i.e., $\log_2 m=2$).

See Supplementary for more experimental results.

\section{Conclusion}
We have proposed a novel two-step training method for MSDA task, namely bi-level optimization based robust target training (\alg{}). \alg{} first learns a labeling function using both the source and target data, then trains a noise-robust model only on the pseudo-labeled target domain. The noise-robust model exploits feature uncertainty to detect label noise and alleviate its negative impact. We further 
employ a bi-level optimization method to optimize the labeling function for better label quality. Extensive experiments on three MSDA datasets demonstrate that our \alg{} achieves new state-of-the-art performance.

\bibliography{egbib}

\newpage

\appendix
\appendixpage

\renewcommand{\appendixname}{~\Alph{section}}
\setcounter{table}{0}
\setcounter{figure}{0}
\setcounter{equation}{0}
\renewcommand{\thetable}{\Alph{table}}
\renewcommand{\thefigure}{\Alph{figure}}
\renewcommand{\theequation}{\Alph{equation}}

This Supplementary Material is organized as follows: Section~\ref{sec:exp_setting} details the experimental settings. 
Section~\ref{sec:fixmatch_cm} introduce the FixMatch-CM. Section~\ref{sec:ablation_fixmatch_cm} presents the ablation study of the FixMatch-CM on PACS. Section~\ref{sec:extra_training_cost} analyzes the extra training cost of our \alg{}. Section~\ref{sec:outer_loop} provides further analysis on the outer loop optimization. 
Section~\ref{sec:cmp_threshold} evaluates the design of adaptive threshold.

\section{Experimental Settings}
\label{sec:exp_setting}
\subsection{Datasets and Protocols} 
We validate the efficacy of our proposed method on three popular MSDA datasets, namely PACS~\cite{li2017deeper}, Digit-Five, and  DomainNet~\cite{Peng_2019_ICCV}. \textbf{PACS} has four different domains (Cartoon, Photo, Sketch and Art Painting), including 9,991 images of 7 categories. We adopt the official train-val splits in~\cite{li2017deeper}.
\textbf{Digit-Five} has five domains, MNIST~\cite{1998Gradient}, SVHN~\cite{37648}, USPS, Synthetic Digits~\cite{syn_digits}, and MNIST-M~\cite{syn_digits}. We follow the protocol in~\cite{Peng_2019_ICCV}. When USPS is used as a source domain, we use all its 9,298 images for training. For the other domains, the training set comprises 25,000 randomly sampled images while the test set has 9,000 images. \textbf{DomainNet} is the largest MSDA dataset available, with about 0.6 million images of 345 categories. These images are collected from six domains, including Sketch, Quickdraw, Painting, Infograph, Real and Clipart. Due to the diversity between different domains in terms of image style, background etc., DomainNet is also the most challenging MSDA dataset so far. 
In all setups, we conduct the leave one held-out protocol and report the average results of three runs.

\subsection{Implementation Details} 
For the first-step labeling function learning in \alg{}, we use  FixMatch-CM (see Section~\ref{sec:fixmatch_cm}) as the model unless stated otherwise. The other details are as follows:
On Digit-Five, we use the backbone with three convolution layers and two fully connected layers, also the same as ~\cite{Peng_2019_ICCV}. The model is optimized with SGD for 30 epochs with initial learning rate 0.05, decayed using the cosine annealing strategy~\cite{loshchilov2016sgdr}, and batch size $B=64$ per domain.
On PACS, we adopt an ImageNet pretrained ResNet-18~\cite{Deep2016He_resnet} as our backbone and optimize it for 100 epochs with Adam~\cite{kingma2014adam}. We set the batch size 16 and the initial learning rate 5e-4.
On DomainNet, an ImageNet pretrained ResNet-101~\cite{Deep2016He_resnet} is used following~\cite{Peng_2019_ICCV}. Then the model is trained with SGD for 40 epochs. The initial learning rate is 0.002. We use batch size $6$ for each domain.

For training the final model in \alg{}, the pseudo-labels are generated in the same way as~\cite{sohn2020fixmatch}. 
The weight of entropy maximization loss $\lambda$ in Eq.~\eqref{eq:loss_inner} is 0.1 and the threshold $m$ in $\mathcal{L}_{ment}$ is 4. The model performance is found to be insensitive to both hyper-parameters (see Section 4.2 in the main paper). We adopt an adaptive threshold for the $\tau$ in Eq. ~\eqref{eq:loss_inner} to filter pseudo labels. Specifically, $\tau$ is initialized with the mean $p_{mean}$ and standard deviation $p_{std}$ of the prediction in a mini-batch, i.e., $p_{mean}+p_{std}$. Then it is gradually decreased to $p_{mean}-p_{std}$ in an exponential moving average way: $\tau = \alpha \tau + (1 - \alpha) (p_{mean}-p_{std})$, where $\alpha$ is fixed to 0.999.  This adaptive scheme is evaluated in Section~\ref{sec:cmp_threshold} and can be regarded as a curriculum sampling strategy.

We initialize the noisy-robust final model  $M_{\psi}$ first by copying the first-step trained MSDA model $F_{\theta}$, followed by adding a stochastic layer in the fourth residual block. 
We start the bi-level optimization when noise-robust model $M_{\psi}$ converges, with a learning rate 5e-5 for fine-tuning the labeling function $F_{\theta}$.

We run our experiments on PyTorch~\cite{paszke2017pytorch,paszke2019pytorch}. Our code is based on Dassl~\cite{paszke2017pytorch,zhou2020domain} \footnote{\url{https://github.com/KaiyangZhou/Dassl.pytorch}}.

\section{FixMatch-CM}
\label{sec:fixmatch_cm}
In this section, we introduce  FixMatch-CM for the first-step MSDA model training. FixMatch-CM adapts the vanilla FixMatch~\cite{sohn2020fixmatch}, originally proposed for semi-supervised learning, to MSDA. Specifically,  FixMatch-CM incorporates two different strong augmentations, image level CutMix~\cite{yun2019cutmix} and feature level MixStyle~\cite{zhou2021domain}, to the vanilla FixMatch for alleviating domain shift further.

Given a batch of source and target images $[ \{(x_i^{\mathcal{S}_1}, y_i^{\mathcal{S}_1}), ...,  (x_i^{\mathcal{S}_K}, y_i^{\mathcal{S}_K})\}_{i=1}^B, \{x_i^{\mathcal{T}}\}_{i=1}^B]$, we first obtain the pseudo-labels $\{\hat{y}_i^{\mathcal{T}}\}_{i=1}^B$ for target images as Eq. (2). Then, following CutMix~\cite{yun2019cutmix}, we crop a patch of random size from each image $x_i^{\mathcal{S}_k}$, and fill in that region with a patch from another randomly sampled image $x^{\mathcal{D}_{k,i}}$ from $\mathcal{D}=\{\mathcal{S}_1, ..., \mathcal{S}_K, \mathcal{T}\}$. Correspondingly, we mix their labels $y_i^{\mathcal{S}_k},y^{\mathcal{D}_{k,i}}$, so we have $\{(\dot{x}_i^{\mathcal{S}_1}, y_i^{\mathcal{S}_1}, y^{\mathcal{D}_{1,i}}, \lambda_i^{\mathcal{S}_1}), ..., (\dot{x}_i^{\mathcal{S}_K}, y_i^{\mathcal{S}_K}, y^{\mathcal{D}_{K,i}}, \lambda_i^{\mathcal{S}_K})\}_{i=1}^B$, $\{(\dot{x}_i^{\mathcal{T}}, \hat{y}_i^{\mathcal{T}}, y^{\mathcal{D}_{T,i}}, \lambda_i^{\mathcal{T}})\}_{i=1}^B$, $\quad$ where $\dot{x}_i^{\mathcal{S}_k}$ is the augmented image, and $\lambda_i^{\mathcal{S}_k}$ is its mixing ratio. 
We also employ a feature-level augmentation in our FixMatch-CM, namely MixStyle~\cite{zhou2021domain}. 
\cite{zhou2021domain} claimed that the mean and standard deviation of $\dot{x}_i^{\mathcal{S}_k}$'s feature map encode domain-specific style statistics. Mixing styles of $\dot{x}_i^{\mathcal{S}_k}$ and $\dot{x}_i^{\bar{\mathcal{D}}(\mathcal{S}_k)}$ can generate interpolated styles, thus augmenting the vanilla feature space, where $\bar{\mathcal{D}}(\mathcal{S}_k)$ are the domains excluding $\mathcal{S}_k$. Then, the style-mixed images $\ddot{x}_i^{\mathcal{S}_k}$ is further forwarded to obtain the prediction $p_i^{\mathcal{S}_k}$. The same operation is also applied to the target images for the prediction  $p_i^{\mathcal{T}}$. Finally, we exploit cross-entropy loss for 
the FixMatch-CM learning:
\begin{equation}
\begin{aligned}
\label{eq:label_fun}
    \underset{\theta}{\arg\min} \ L(\theta) & =\frac{1}{KB}\sum_k^K\sum_i^B [\lambda_i^{\mathcal{S}_k} L_{ce}(p_i^{\mathcal{S}_k}, y_i^{\mathcal{S}_k}) \\ 
    & \quad + (1 - \lambda_i^{\mathcal{S}_k}) L_{ce}(p_i^{\mathcal{S}_k}, y^{\mathcal{D}_{k, i}})]\\
    & + \frac{1}{B}\sum_i^B [\mathbbm{1}(q(\hat{y}_i^{\mathcal{T}}) \geq \tau_0) \lambda_i^{\mathcal{T}} L_{ce}(p_i^{\mathcal{T}}, \hat{y}_i^{\mathcal{T}})\\
    & \quad + (1-\lambda_i^{\mathcal{T}}) L_{ce}(p_i^{\mathcal{T}}, y^{\mathcal{D}_{T,i}})],
\end{aligned}
\end{equation}
where $q(\hat{y}_i^{\mathcal{T}})$ is the predicted probability corresponding to $\hat{y}_i^{\mathcal{T}}$. $\tau_0$ is a threshold to filter out low-confidence pseudo-labels. It is fixed as 0.95 same as FixMatch~\cite{sohn2020fixmatch}. We focus on this first-step MSDA training method in our experiments.

\subsection{Ablation Study of FixMatch-CM} 
\label{sec:ablation_fixmatch_cm}
Figure ~\ref{fig:other_sota_with_NTCT} of the main paper shows that FixMatch repurposed for DA achieves the state of the art performance on PACS benchmark already. Our FixMatch-CM further improves it.
 To better understand FixMatch-CM, we investigate its each component. First, we discard the image-level strong augmentation -- CutMix~\cite{yun2019cutmix} and observe an 0.75\% accuracy drop from 93.89\% to 93.14\% (c.f. \#\ref{ablmodel:fm_da} vs. \#\ref{ablmodel:fm_ms_w_dm} in Table~\ref{tab:ablation_fixmatch_da}). Further removing MixStyle~\cite{zhou2021domain} from \#\ref{ablmodel:fm_ms_w_dm} leads to a 4.35\% accuracy drop as we assume that vanilla FixMatch can only weakly deal with the domain shift.

\begin{table}[tb]
    \centering
    \caption{Ablation study on FixMatch-CM.}
    \begin{tabular}{l|l|cccc|c}
    \hline
    \textbf{\#} &\textbf{Methods} & \textbf{Avg}\\
    \hline
    \ablmodel  \label{ablmodel:fm_da} & FixMatch-CM (MixStyle + CutMix) &93.89\\
    \ablmodel  \label{ablmodel:fm_ms_w_dm}  & FixMatch (MixStyle only)  & 93.14\\
    \ablmodel  \label{ablmodel:fm_vanilla} & FixMatch (vanilla) &88.79 \\
    \ablmodel  \label{ablmodel:src_only} & Source only  &82.17 \\
    \hline
    \end{tabular}
    \label{tab:ablation_fixmatch_da}
\end{table}

\section{Extra Training Cost of \alg{}}
\label{sec:extra_training_cost}
The major limitation of our \alg{} is the extra training cost brought by the second-step training. To reduce the training cost, we further evaluate our \alg{} on PACS by controlling the total training epochs.
We keep the total training epoch exactly the same as one-step training MSDA methods, i.e., 100 epochs (50 epochs for the first-step labeling function training and the rest for the second-step target model training). From the results in Table~\ref{tab:training_cost}, we can see that reducing the total training epochs makes the performance of our \alg{} worse, however, it still clearly outperforms the base method FixMatch-CM by $\sim0.7\%$. These results not only show the efficacy of our proposed \alg{} and demonstrate that our \alg{} works even if the labeling function is not thoroughly trained in the first step.

\begin{table}[tb]
    \centering
    \caption{Reducing the training cost for \alg{}.}
    \begin{tabular}{l|cccc|c}
    \hline
    \textbf{Methods} & \textbf{Avg}\\
    \hline
    \alg{} &95.38\\
    \alg{} (Control training epochs)  & 94.57\\
    FixMatch-CM & 93.89\\
    \hline
    \end{tabular}

    \label{tab:training_cost}
\end{table}

\section{Further Analysis on Outer Loop Optimization}
\label{sec:outer_loop}
In Eq.~\eqref{eq:loss_outer} of the main paper, we minimize the entropy (or feature uncertainty) loss on current training set for the outer loop. Here we further conduct experiments to see 1) whether the current training set is better than a held-out validation set, and 2) whether the feature uncertainty loss as objective is better than pseudo-label based cross-entropy loss. We show the comparative results in Table~\ref{tab:train_vs_val}. We can see from the first two rows that minimizing feature uncertainty loss on a held-out validation set obtains slightly worse performance on PACS and DomainNet. This observation suggests that a held-out validation set is not necessary for the feature uncertainty minimization objective. When we use such validation set to calculate cross-entropy for the outer loop, the performance on DomainNet even decreases from 53.4\% to 52.7\%. We assume that the degradation is caused by the noise in the pseudo-labels. Overall, the entropy loss optimized on current training batch achieves the best performance over the other alternatives, and saves the labor for splitting a held-out validation set.
\begin{table}[t]
    \centering
    \tabstyle{6pt}
    \caption{Comparison of using training set and held-out validation set for the outer loop optimization. Entropy loss denotes the Eq.(7) in main paper.}
    \begin{tabular}{c|ccc}
        \hline
        \textbf{Setting} & \textbf{Digit-Five} & \textbf{PACS} & \textbf{DomainNet} \\
        \hline
        Feature uncertainty loss on training set & 95.9 & 95.38 & 53.4\\
        Feature uncertainty loss on validation set & 95.9 & 95.19 &53.2\\
        Cross-entropy on validation set & 95.8 & 95.20 & 52.7\\
        \hline
    \end{tabular}
    \label{tab:train_vs_val}
\end{table}

\section{Ablation Study on the Adaptive Threshold}
\label{sec:cmp_threshold}
\begin{table}[tb]
    \centering
    \caption{Ablation study on adaptive threshold.}
    \begin{tabular}{l|c}
    \hline
    \textbf{Methods} & \textbf{Avg}\\
    \hline
    \alg{} (fixed $\tau=0.95$) & 95.21 \\
    \alg{} (adaptive $\tau$ for each class) & 95.36 \\
    \alg{} (adaptive $\tau$) & 95.38 \\
    \hline
    \end{tabular}
    \label{tab:adaptive_threshold}
\end{table}
We practically adopt an adaptive threshold in Eq.~\eqref{eq:loss_inner} as an alternative to fixed threshold, e.g., $\tau=0.95$ as in FixMatch~\cite{sohn2020fixmatch}. 
We compare the adaptive threshold with the fixed threshold in Table~\ref{tab:adaptive_threshold}. We can observe that the adaptive threshold works better than the fixed threshold. This is possibly because the adaptive threshold gradually includes more and more samples for training in an easy-to-hard way, leading to a curriculum learning strategy. Moreover, we found that different MSDA datasets usually need different $\tau$ for optimal model performance, e.g., fixing $\tau$ to 0.95 works well on PACS but causes poor performance on DomainNet (accuracy$\approx$36\%).
Therefore, with this adaptive threshold a hyper parameter tuning is perfectly saved.

Considering that class imbalance can happen, we also try using different thresholds for different classes, i.e., for each class $c$, we set an adaptive threshold $\tau_c$ for that class. Here, $\tau_c$ is updated via $\tau_c = \alpha \tau_c + (1 - \alpha) (p^c_{mean}-p^c_{std})$, with  $p^c_{mean}$ and $p^c_{std}$ denoting the mean and standard deviation of the predictions of class $c$ in a mini-batch.  This alternative choice achieves 95.36\% on PACS, similar to using a single threshold for all the classes (95.38\%). This is probably because a single threshold can already pick a reasonable amount of samples in each class on PACS.

\begin{figure*}[t]
    \centering
    \subfigure[M$^3$SDA]{
        \begin{minipage}[t]{0.4\textwidth}
        \includegraphics[trim=60 60 60 65,clip,width=\columnwidth]{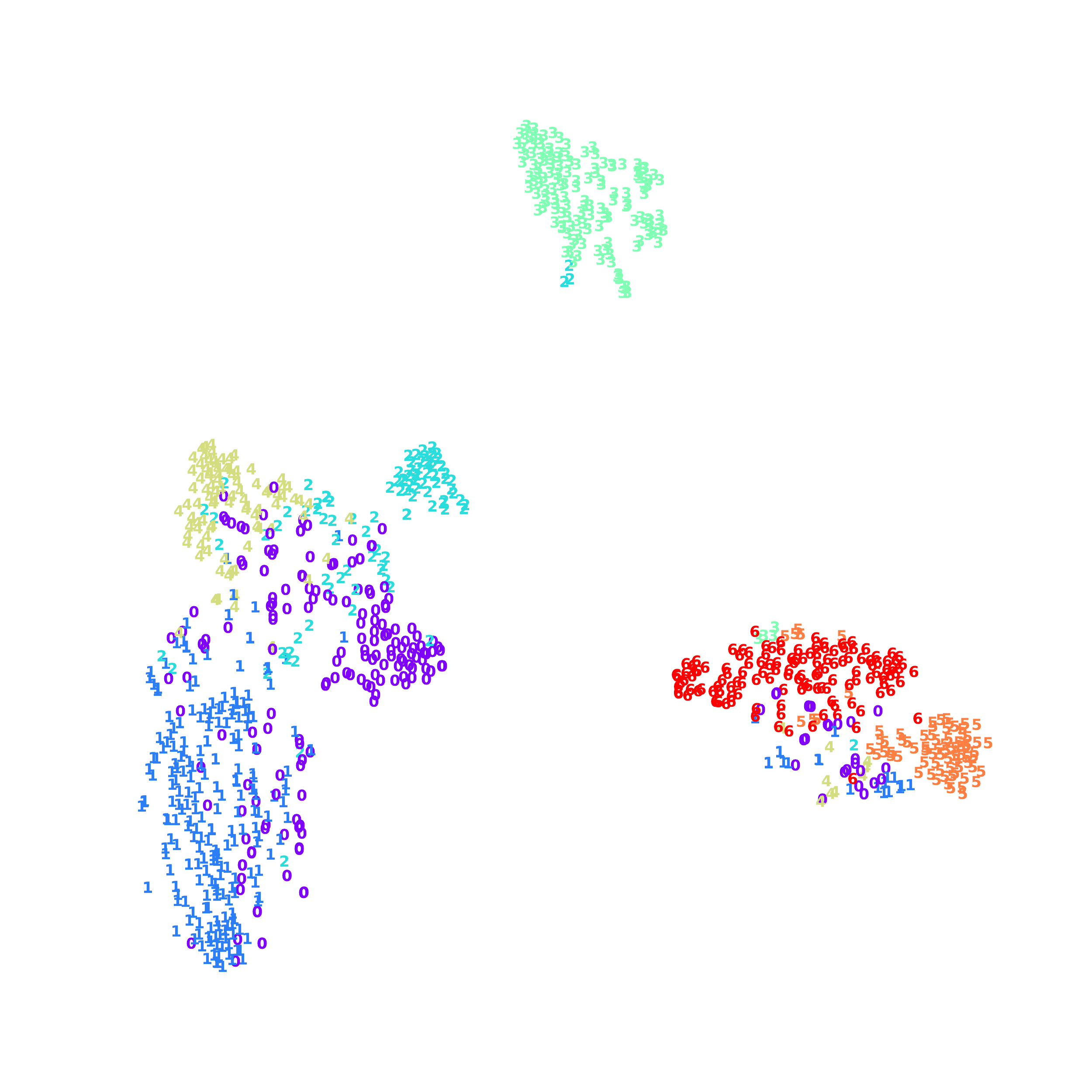}
        \label{fig:m3sda}
        \end{minipage}
    }
    ~\hfill
    \subfigure[DRT]{
        \begin{minipage}[t]{0.4\textwidth}
        \includegraphics[trim=60 60 60 65,clip,width=\columnwidth]{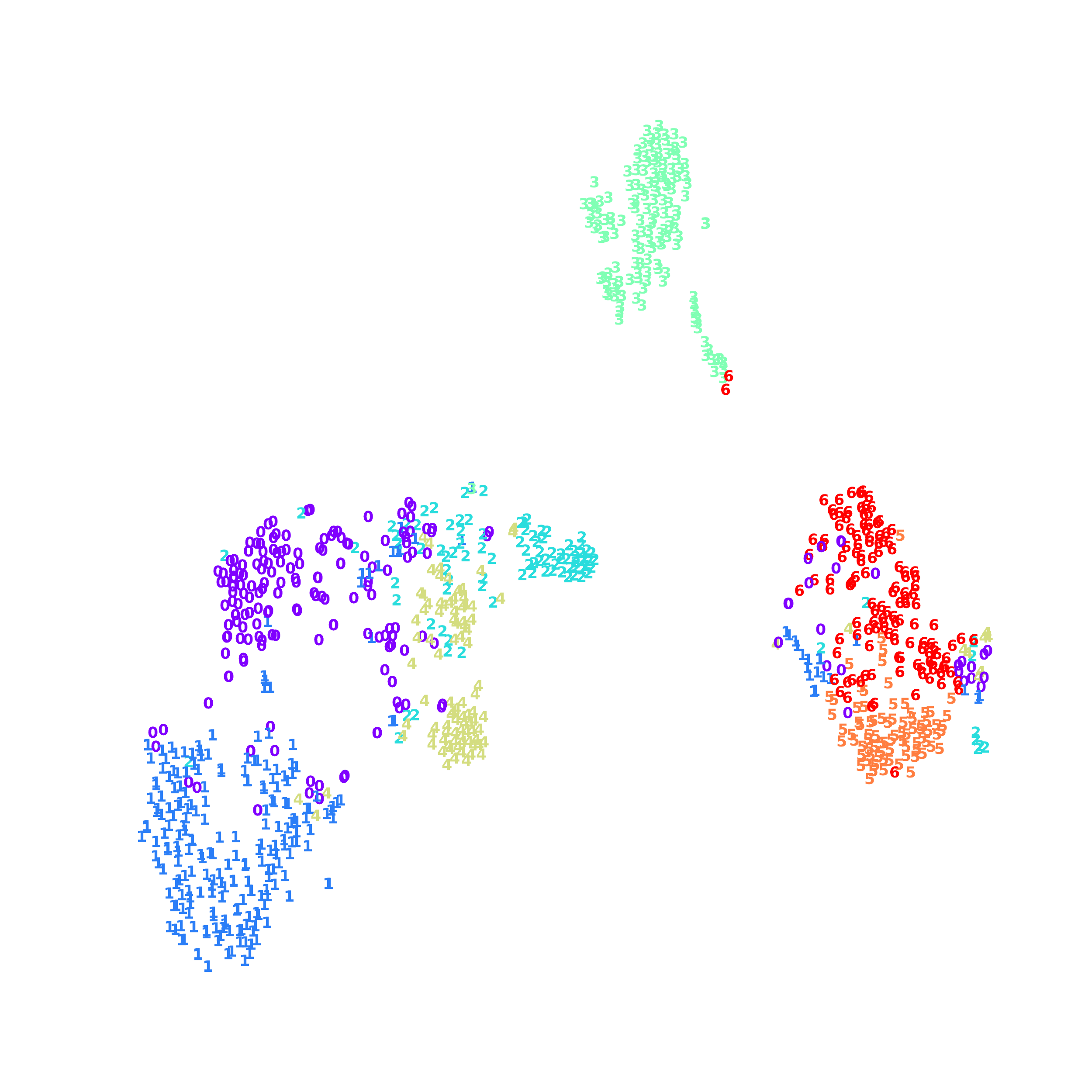}
        \label{fig:drt}
        \end{minipage}
    }
    ~\hfill
    \subfigure[FixMatch-CM]{
       \begin{minipage}[t]{0.4\textwidth}
        \includegraphics[trim=60 60 60 65,clip,width=\columnwidth]{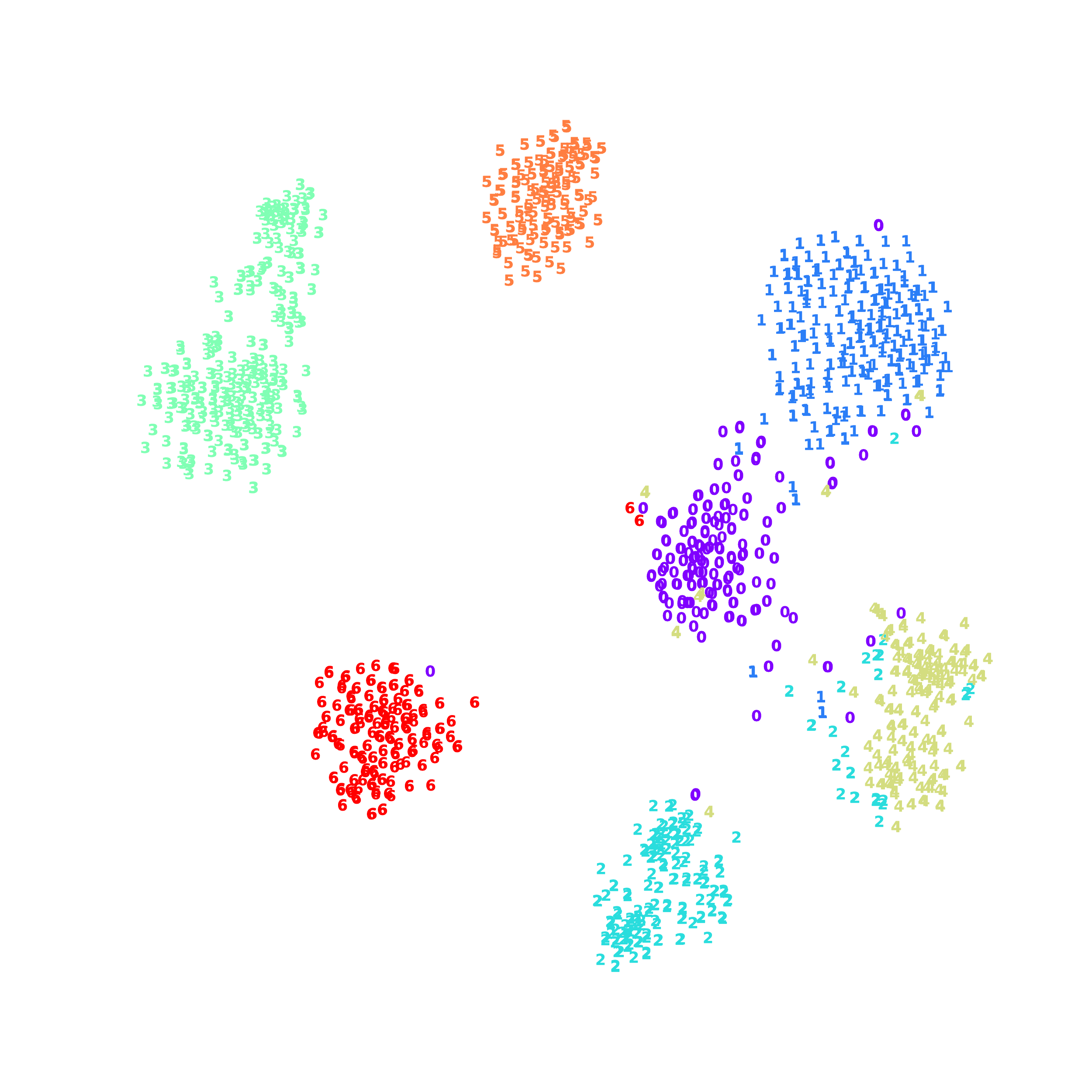}
        \label{fig:fixmatch_cm}
        \end{minipage}
    }
    ~\hfill
    \subfigure[\alg{}]{
        \begin{minipage}[t]{0.4\textwidth}
        \includegraphics[trim=70 70 70 70, clip, width=\columnwidth]{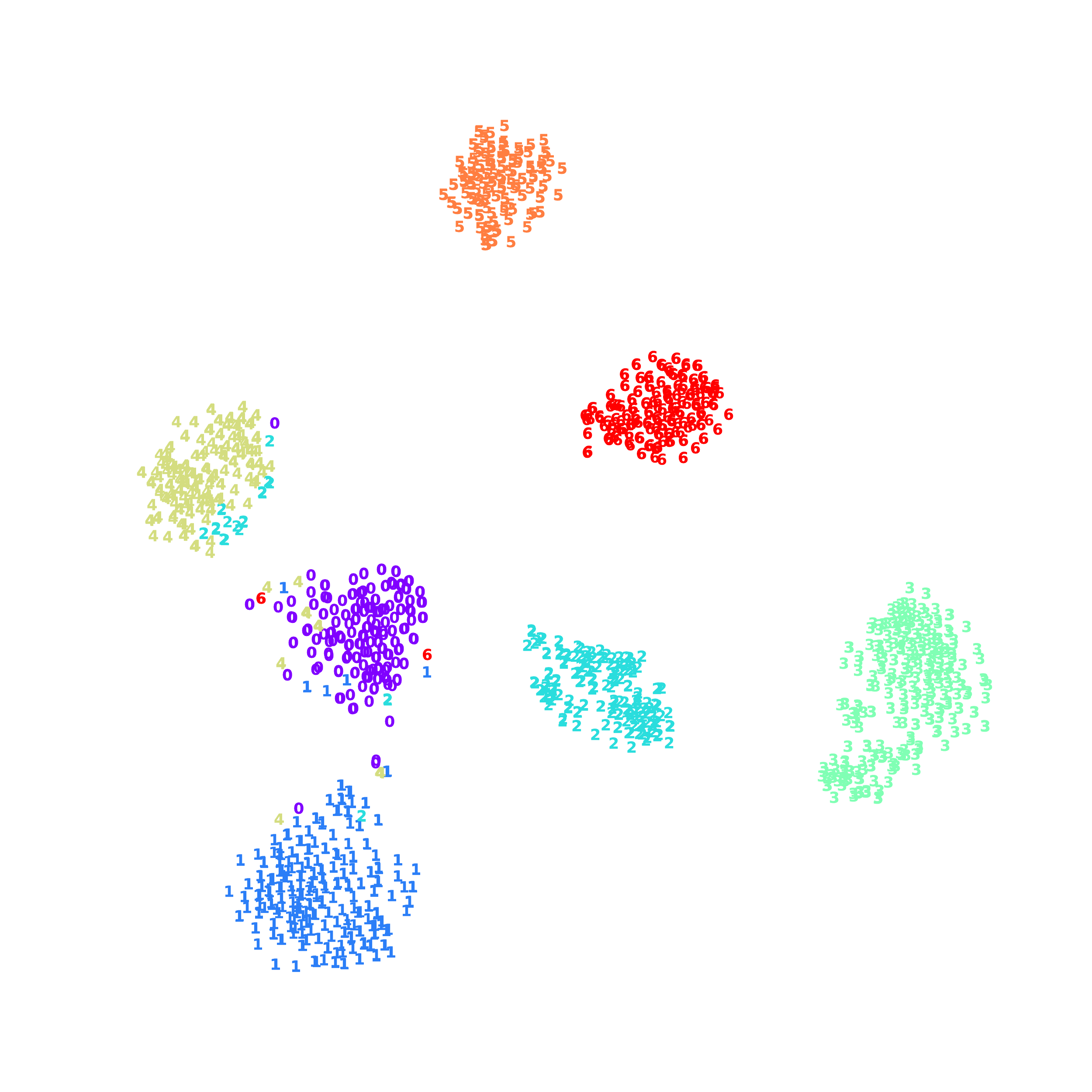}
        \label{fig:ntct}
        \end{minipage}
    }
    \caption{Visualization of features from M$^3$SDA, DRT, FixMatch-CM and \alg{} on PACS (target domain: Sketch) using t-SNE~\cite{maaten2008visualizing}. Different colors (or digits 0-6) denote different classes. 
    Better viewed with zoom-in.}
    \label{fig:visualization_ntct}
\end{figure*}

\section{Visualization of Learned Features}

To better understand how our \alg{} works, we further provide a t-SNE visualization of feature distributions in Figure~\ref{fig:visualization_ntct}. 
From Figure~\ref{fig:visualization_ntct}, we can see that FixMatch-CM enables the better feature separability compared with M$^3$SDA and DRT, which illustrates the effectiveness of FixMatch-CM. Based on FixMatch-CM, our \alg{} further increases its inter-class distance, leading to the best class-wise separability. We attribute this to our two-step training pipeline which eliminates the source domain bias in our model.
\end{document}